\documentclass[sigconf]{acmart}
\AtBeginDocument{%
  }

\copyrightyear{2026}
\acmYear{2026}
\setcopyright{cc}
\setcctype{by}
\acmConference[CIKM '26]{Proceedings of the 35th ACM International Conference on Information and Knowledge Management}{November 07--11, 2026}{Rome, Italy}
\acmBooktitle{Proceedings of the 35th ACM International Conference on Information and Knowledge Management (CIKM '26), November 07--11, 2026, Rome, Italy}
\acmDOI{10.1145/3799682.3839901}
\acmISBN{979-8-4007-2539-5/2026/11}
\begin{document}

\title[Pair-Level Essay-Scale Republication and Reuse]{Pair-Level Essay-Scale Republication and Reuse from Fragmented Historical Text Reuse: A Workflow Study on Eighteenth-Century Books and Newspapers}

\author{Ke Shu}
\orcid{0009-0008-3574-1630}
\email{ke.shu@helsinki.fi}
\affiliation{%
  \institution{University of Helsinki}
  \city{Helsinki}
  \country{Finland}
}
\author{Kira Hinderks}
\orcid{0009-0004-9233-9315}
\email{kira.hinderks@helsinki.fi}
\affiliation{%
  \institution{University of Helsinki}
  \city{Helsinki}
  \country{Finland}
}
\author{Eetu Mäkelä}
\orcid{0000-0002-8366-8414}
\email{eetu.makela@helsinki.fi}
\affiliation{%
  \institution{University of Helsinki}
  \city{Helsinki}
  \country{Finland}
}
\author{Mikko Tolonen}
\orcid{0000-0003-2892-8911}
\email{mikko.tolonen@helsinki.fi}
\affiliation{%
  \institution{University of Helsinki}
  \city{Helsinki}
  \country{Finland}
}

\begin{abstract}
This paper addresses the recovery of essay-scale republication and reuse from fragmented text-reuse evidence, a setting whose central challenge is pair-level evidence consolidation and not fragment retrieval alone. The study focuses on a candidate set centered on essays by eighteenth-century Scottish philosopher David Hume, spanning books from ECCO (Eighteenth Century Collections Online) and historical newspapers. Because the input consists of fragmented reuse hits instead of clean document pairs, and positive coverage is inherently incomplete, we formulate the task as pair-level evidence consolidation into plausible transmission relations and compare three methodological families: a staged rule-based workflow, baselines (a decision tree and two direct LLM settings), and automated rule adaptation. On labeled ECCO--ECCO slices, pair-level feature aggregation alone already reaches 0.948 F1 on the main labeled slice, while the final workflow gives the strongest overall precision-recall trade-off among the tested rule stages. On the full ECCO--ECCO candidate universe, direct LLM baselines flag up to 14,886 pairs as reprints compared to 771 for the final workflow, behaving in this direct-prompt setup as high-recall candidate expanders rather than precision-controlled deployment classifiers. On ECCO--Newspaper, manual audit confirms all 176 predicted positives as genuine cases of republication or reuse, while issue duplication and source-side multiplicity reveal additional provenance structure. Under incomplete ground truth, auditable pair-level evidence consolidation provides a practical way to produce compact candidate spaces for historical inspection.
\end{abstract}



\maketitle
\section{Introduction}
This paper asks how historians can recover essay-scale republication and reuse across eighteenth-century books and newspapers from fragmented text-reuse evidence. Historians of the Enlightenment have documented the wide circulation of ideas across books, pamphlets, and newspapers, often without explicit attribution~\cite{darnton1982history}, leaving transmission chains that are difficult to reconstruct from catalog metadata alone. We therefore focus on recovering substantial reused argumentative units that can support downstream historical case studies. Bridging two structurally different historical environments is technically demanding: books carry a stable bibliographic identity, whereas in newspapers text circulates through excerpt, recomposition, and redistribution, so textual identity itself becomes unstable. Prior systems have addressed book or newspaper reuse in isolation, not their systematic crossing.

The emphasis on \emph{essay-scale} is methodological as well as historical. We are not only interested in short overlaps or perfectly identical copies. In our material, genuine cases of reuse may involve paragraph reordering, abridgment, local rewriting, and different page or column segmentation across media. We therefore adopt an operational definition centered on Hume's essays as known source texts: if cross-media transmission preserves a substantial portion of the source essay---operationalized through coverage and span features---we treat the pair as an instance of essay-scale republication or reuse even when local wording or layout changes. Hume is a historically motivated seed case: his works were widely reprinted in the period and are supported by well-documented metadata and prior scholarship, though direct portability to other authors remains to be tested.

The task is better understood as consolidating fragmented evidence into assessable transmission relations than as supervised classification: the input is a noisy, high-recall set of fragment-level reuse hits, and the key question is whether those local matches jointly support a relation of substantial essay-scale reuse. Pair-level evidence consolidation, not fragment retrieval, is therefore the central problem. It also means that extensive textual overlap does not automatically imply reprint, and that full-corpus evaluation is complicated by incomplete positive coverage.

Our contribution is a workflow-centered study over 17 Hume books from ECCO and two target domains: ECCO books and the Burney Newspapers Collection. We compare three methodological families built on the same pair-level evidence space: a staged rule-based workflow, three concrete baselines (a decision tree and two direct LLM settings), and automated rule adaptation. The three families represent increasing degrees of automation over the same evidence space.

This framing builds on four adjacent lines of work. Historical text-reuse research has already shown that large-scale reprint and reuse patterns can be recovered from noisy corpora. The \emph{Viral Texts} project and the Passim tool that underlies it~\cite{smith2015computational} introduced local text-reuse detection for historical newspapers; \emph{Reception Reader}~\cite{rosson2023reception} extended this into an interactive exploration interface. However, these systems mainly emphasize retrieval and exploration rather than pair-level evidence consolidation into transmission relations. From a learning perspective, our setting is closer to positive-focused or incompletely labeled inference than to fully observed supervised classification \cite{bekker2020learning,buckley2004retrieval}. It is also closely related to weak supervision and labeling systems such as Snorkel and Errudite, and to later interactive or LLM-assisted approaches that iteratively induce, discover, or apply rules and prompts as labeling signals \cite{ratner2017snorkel,wu2019errudite,boecking2020interactive,galhotra2021adaptive,smith2024language}. Our use of page evidence is informed by layout-aware document understanding~\cite{xu2020layoutlm,appalaraju2021docformer,gutehrle2022processing}, though here layout serves as targeted support for hard-case refinement rather than end-to-end classification input.

We make three contributions: (1)~a pair-level formulation of essay-scale republication and reuse detection that extends fragment-level retrieval systems such as Passim~\cite{smith2015computational} with evidence consolidation under incomplete positive coverage, enabling transmission-relation judgment rather than fragment retrieval alone; (2)~a comparative evaluation across two historically distinct environments, ECCO books and eighteenth-century newspapers, a crossing not previously studied at essay scale, showing that staged workflow design can produce precision-controlled candidate spaces historians can efficiently evaluate; and (3)~empirical evidence that under incomplete positive coverage, closed-world F1 alone is insufficient, because evidence consolidation, deployment behavior, and historian inspection each contribute distinct quality signals.

\section{Task, Data, and Methods}
We construct a pair-level candidate universe from fragment-level reuse hits produced by BLAST-style local sequence alignment~\cite{smith2015computational}, where each hit records a short locally-aligned passage shared between a source and a target document. This fragment-matching step supplies input evidence; our naive baseline is therefore not a comparison against the full Passim system, but a shallow pair-level rule applied after fragment retrieval. The system must aggregate those hits into pair-level evidence and decide whether the pair constitutes an instance of essay-scale republication or reuse. The shared feature space includes coverage, span, section distribution, fragment chaining, and cue-like signals such as title, heading, quotation, and paratext indicators.

Our source subset is centered on 17 books by David Hume (1711--1776), the Scottish philosopher whose essays circulated widely in eighteenth-century print culture. The study design grew from the observation that even naive fragment-level reuse signals reduce the candidate space from millions of potential document pairs across ECCO to tens of thousands, a contraction that already enables efficient expert curation and makes candidate-space control a central part of discovery quality. We study two target domains: books from ECCO (Eighteenth Century Collections Online) and eighteenth-century newspapers from the Burney Newspapers Collection~\cite{burney_collection}. To avoid trivial within-author carryover, we explicitly filter out straightforward Hume-to-Hume internal reprints among later editions, so the task concentrates on Hume-to-others reuse and reprint relations rather than on edition matching inside Hume's own corpus. The ECCO--ECCO candidate universe contains 22{,}735 pairs, including 369 labeled evaluation pairs with 111 known positives (60 in \texttt{main}, 22 in \texttt{hard}). The ECCO--Newspaper universe contains 3{,}583 pairs. On the newspaper side, we treat the data as a discovery setting: we first run the final workflow over the candidate space, then manually validate the predicted positives and audit a smaller sample of predicted negatives.

The key modeling move is to shift attention from individual reuse hits to locally consolidated pair-level evidence. In practice, this means that several nearby fragments may jointly support one reused passage even when no single hit is decisive on its own. Features such as bundle span, cumulative coverage, section concentration, and chaining strength therefore act as proxies for whether fragmented evidence is cohering into a larger discursive unit instead of only reflecting scattered quotation or diffuse borrowing. The workflow therefore aggregates fragment-level signals into a pair-level case before any reuse judgment is made, so the unit of interpretation is the evidence bundle and not the individual fragment.

\begin{figure}[b]
\centering
\includegraphics{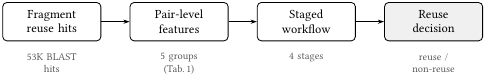}
\caption{Overview pipeline: fragment hits are aggregated into pair-level features, then classified by the staged rule workflow.}
\label{fig:pipeline}
\end{figure}

We use \emph{workflow} in a narrow sense: the staged rule-based pair classifier applied after fragment retrieval and pair-level feature extraction. The broader discovery process includes fragment generation, feature aggregation, workflow stages/ablations, baselines, automated adaptation, and manual audit; the workflow itself refers only to the explicit rule cascade. Table~\ref{tab:features} summarizes the five feature groups shared across all methods, and Figure~\ref{fig:pipeline} shows the overall pipeline from fragment hits to reuse decision.

\begin{table}[t]
\centering
\small
\setlength{\tabcolsep}{4pt}
\caption{Pair-level features used across all methods.}
\label{tab:features}
\begin{tabular}{@{}lp{4.6cm}@{}}
\toprule
Feature group & Description \\
\midrule
Coverage       & Fraction of source essay spanned by reuse hits \\
Span / bundle  & Max and cumulative span of contiguous hit bundles \\
Fragment chain & Longest monotone hit chain ordered by source offset \\
Section conc.  & Hit concentration across document sections \\
Cue signals    & Title, heading, quotation, paratext hit counts \\
\bottomrule
\end{tabular}
\end{table}

\paragraph{Rule-based workflow.}
The workflow family is the main methodological line of the paper. It applies four increasingly informative rule stages over the same pair-level feature table: a naive heuristic based on a single shallow span signal, a stronger structural-only stage built from coverage and bundle statistics, a context-aware stage that adds title, heading, quotation, and paratext cues, and the final workflow, which adds hard-case rescue and suppression. The logic is cumulative: pair-level aggregation supplies the main signal, while later stages sharpen the decision boundary around ambiguous reuse. Table~\ref{tab:rule-cascade} exposes the retained final cascade.

\begin{table}[t]
\centering
\scriptsize
\setlength{\tabcolsep}{3pt}
\caption{Final rule cascade used by the retained workflow. Spans are in characters; fanout is the number of distinct source sections strongly linked to the same target document.}
\label{tab:rule-cascade}
\begin{tabular}{@{}p{1.45cm}p{6.1cm}@{}}
\toprule
Gate & Rule logic \\
\midrule
Structural & Accept compact chains with span $\geq$900, coverage $\geq$0.09, sum coverage $\geq$0.10, section count $\leq$8, fanout $\leq$40, and chain score $\geq$1200; or denser bundles with span $\geq$650, coverage/sum coverage $\geq$0.15, section count $\leq$4, fanout $\leq$8, and at least two fragments. \\
Coverage & Also accept high-coverage pairs with coverage $\geq$0.30, fanout $\leq$40, and section count $\leq$8. \\
Context & Require either title/heading evidence or the absence of broad multi-section/multi-fragment dispersion; reject paratext-only and quotation-dominated cases in the non-cued branch. \\
Rescue & If no quotation or paratext cue and destination fanout is zero, rescue short near misses with span 450--870 and either coverage/section-gap or sum-coverage/section-count conditions. \\
\bottomrule
\end{tabular}
\end{table}

The cascade is not a retrained model but a fixed, inspectable decision boundary. Each positive decision can be traced to a structural, coverage, context, or rescue gate, giving reviewers and later users a concrete reason for why the pair entered the candidate set. The rescue branch is deliberately narrow: it only admits low-fanout near misses without quotation or paratext cues, rather than acting as a general recall booster. Conversely, suppression logic does not claim that quotation, paratext, or dispersed matches are never historically meaningful; it keeps them out of the precision-controlled output so that they can be handled by manual or follow-up review.

\paragraph{Baselines.}
The baselines test whether the same evidence space can be exploited without an explicit workflow. The decision tree was trained once on the discovery split (60 pairs, balanced 30/30 positive/negative) and applied unchanged to all other splits; no cross-validation was used. The two direct LLM settings used Qwen3-30B-A3B-Instruct-2507~\cite{qwen3} (greedy decoding, 128-token output limit): a text-only variant providing the top-5 ranked reuse fragments and light pair metadata, and a structured variant that additionally serializes the pair-level feature signals. Both share the same system prompt and a binary reprint/non-reprint output schema. These direct LLM runs are not matched white-box classifiers over fixed features: they may also draw on pretrained knowledge of Hume, titles, metadata patterns, or historically common phrases. We therefore treat them as direct-prompt reference conditions for candidate expansion, not as evidence for a general claim about LLMs for historical reuse discovery. The practical distinction matters: an effective historical workflow does not replace expert judgment but produces a candidate space small enough for efficient historian inspection.

\paragraph{Automated adaptation.}
The third family asks whether the transition from the context-aware stage to the strongest final workflow can be partially automated. Here the system constructs hard-case pools from stage-to-stage differences, induces or proposes bounded candidate rules, and redeploys them over the same pair universe. When an LLM is used in this pipeline, its role is not to act as the final classifier, but to propose candidate rescue or suppression rules under a bounded rule schema. This keeps the adaptive step auditable and lets the same induced rules be validated both on labeled anchors and on the full candidate universe.

Automated adaptation thus provides a reproducible starting point for new corpora where no hand-labeled hard cases yet exist, positioned between the broad LLM baseline and the labor-intensive hand-tuned workflow.

We evaluate at three levels. On labeled ECCO--ECCO slices, we report pair-level precision, recall, and F1 on \texttt{main}, \texttt{hard}, and \texttt{all-labeled}; the \texttt{hard} slice should be read as a diagnostic of boundary behavior rather than a fully held-out generalization test, since the final workflow's rescue and suppression rules were refined using hard-case pools that overlap with this split. On the full ECCO--ECCO universe, we compare deployment behavior through positive-output size and anchor recovery. On ECCO--Newspaper, we report manual audit results.

\section{Results}
Table~\ref{tab:ecco-labeled-comparison} shows the ECCO--ECCO labeled-slice results. The first takeaway is that pair-level aggregation itself is already strong: the structural-only stage reaches 0.948 F1 on \texttt{main}, and even the decision tree reaches 0.909. The real difficulty is concentrated in hard cases, where the decision tree drops to 0.063 F1 and the naive and structural-only workflows remain at 0.171. The final workflow does not dominate every method on every metric, but it gives the strongest overall balance on labeled ECCO--ECCO, reaching 0.825 F1 on \texttt{all-labeled} and the highest \texttt{hard} precision (0.333) of any method. Its \texttt{hard} F1 is not the highest, and that slice in any case overlaps with hard-case pools used in workflow refinement (see Limitations). By contrast, the two direct LLM baselines behave more like high-recall candidate finders than precision-controlled tools for historian inspection.

This comparison also clarifies the role of the three experimental families. The rule-based progression shows where the gains actually come from: hard-slice F1 stays at 0.171 through the naive and structural-only stages and rises to 0.263 once contextual cues are added, while the final stage contributes boundary control rather than raw recovery, lifting hard-slice precision from 0.312 to 0.333 and cutting deployment output from 1{,}265 to 771 positives. The decision tree shows that the feature space itself is informative enough for a strong shallow classifier, but that this learned boundary remains brittle on difficult cases. Automated adaptation sits between the two other families: it recovers far more hard-slice positives than the hand-designed stages (0.455 recall, 0.317 F1, against 0.227 and 0.270 for the final workflow), but at lower hard-slice precision and lower \texttt{all-labeled} F1 (0.776 against 0.825), and with a 56\% larger deployment output.
\begin{table*}[t]
\centering
\caption{Main results on labeled ECCO--ECCO slices. All methods are scored against the same gold labels: 60, 22, and 111 positives in \texttt{main}, \texttt{hard}, and \texttt{all-labeled}.}
\label{tab:ecco-labeled-comparison}
\begin{tabular}{lccccccccc}
\toprule
& \multicolumn{3}{c}{Main} & \multicolumn{3}{c}{Hard} & \multicolumn{3}{c}{All-labeled} \\
\cmidrule(lr){2-4} \cmidrule(lr){5-7} \cmidrule(lr){8-10}
Method & P & R & F1 & P & R & F1 & P & R & F1 \\
\midrule
Naive heuristic & 0.851 & 0.950 & 0.898 & 0.125 & 0.273 & 0.171 & 0.672 & 0.811 & 0.735 \\
Structural-only & 0.982 & 0.917 & 0.948 & 0.231 & 0.136 & 0.171 & 0.890 & 0.730 & 0.802 \\
Context-aware & 0.778 & 0.933 & 0.848 & 0.312 & 0.227 & 0.263 & 0.752 & 0.766 & 0.759 \\
Decision tree & 0.833 & 1.000 & 0.909 & 0.041 & 0.136 & 0.063 & 0.568 & 0.829 & 0.674 \\
LLM text-only & 0.341 & 1.000 & 0.508 & 0.217 & 0.909 & 0.351 & 0.357 & 0.982 & 0.524 \\
LLM structured & 0.395 & 1.000 & 0.566 & 0.202 & 0.773 & 0.321 & 0.394 & 0.937 & 0.555 \\
Automated adaptation & 0.829 & 0.967 & 0.892 & 0.244 & 0.455 & 0.317 & 0.709 & 0.856 & 0.776 \\
Final workflow & 0.919 & 0.950 & 0.934 & 0.333 & 0.227 & 0.270 & 0.870 & 0.784 & 0.825 \\
\bottomrule
\end{tabular}
\end{table*}

Table~\ref{tab:ecco-deploy} shows deployment behavior on the full candidate universe. Direct LLM baselines produce the broadest positive outputs (14{,}886 and 11{,}223) and attain the highest recall on every labeled split, but their unlabeled output would be costly to inspect directly. The decision tree, the context-aware stage, and automated adaptation all settle in the 1.2--1.3K range, well above the naive, structural-only, and final workflow stages. The final workflow maintains the most controlled output (771 positives) while retaining competitive main-slice recall (MR\,=\,0.950); its lower hard-slice recall (HR\,=\,0.227) reflects the same conservative boundary that limits false positives in the unlabeled majority.

\begin{table}[t]
\centering
\setlength{\tabcolsep}{4pt}
\caption{Deployment on the full ECCO--ECCO universe. \#Pos = predicted positive count, restricted to the 22{,}735 shared pair universe; AR/MR/HR = recall on the \texttt{all-labeled}, \texttt{main}, and \texttt{hard} anchor sets.}
\label{tab:ecco-deploy}
\begin{tabular}{lrrrr}
\toprule
Method & \#Pos & AR & MR & HR \\
\midrule
Naive heuristic  & 961    & 0.811 & 0.950 & 0.273 \\
Structural-only  & 710    & 0.730 & 0.917 & 0.136 \\
Context-aware    & 1{,}265  & 0.766 & 0.933 & 0.227 \\
Decision tree    & 1{,}242  & 0.829 & 1.000 & 0.136 \\
LLM text-only    & 14{,}886 & 0.982 & 1.000 & 0.909 \\
LLM structured   & 11{,}223 & 0.937 & 1.000 & 0.773 \\
Automated adaptation & 1{,}206  & 0.856 & 0.967 & 0.455 \\
Final workflow   & 771    & 0.784 & 0.950 & 0.227 \\
\bottomrule
\end{tabular}
\end{table}

The ECCO--Newspaper setting is not simply a second test domain: unlike ECCO--ECCO, where partially labeled pairs support quantitative evaluation, the newspaper side involves a structurally different historical environment with no pre-existing ground truth, requiring a deployment-and-audit approach instead of closed-world measurement. Manual annotation assigned each audited pair a reprint, non-reprint, or uncertain label, with an optional free-text note recording the annotator's reason. Notes were kept only for boundary cases and do not form a coded taxonomy. One primary annotator carried out the checks; hard or boundary cases were discussed with two further domain experts as targeted adjudication, not independent double coding, so no inter-annotator agreement is reported.

All 176 newspaper cases predicted as republication or reuse by the final workflow were manually confirmed as genuine. We also audited 49 predicted negatives selected for diagnostic inspection, including near-boundary and disagreement cases, of which 46 were resolved into 38 non-reprints and 8 reprints. Because this negative audit intentionally oversamples difficult cases, its false-negative rate should not be read as population prevalence. On the 222 resolved audit cases (176 predicted positives + 46 resolved negatives), precision, recall, and F1 are 1.000, 0.957, and 0.978; these figures reflect the audited subset, not corpus-level recall over all 3{,}583 newspaper pairs. Attribution was coded for 175 of the predicted positives; 127 of these name Hume explicitly and 48 do not, indicating that the workflow recovers both attributed and unattributed transmission.

\begin{table}[t]
\centering
\caption{Manual audit of ECCO--Newspaper results.}
\label{tab:newspaper-audit}
\begin{tabular}{lr}
\toprule
Item & Value \\
\midrule
Predicted positives (fully annotated) & 176, all genuine \\
Predicted negatives (audited) & 49 \\
Resolved as non-reprint & 38 \\
Resolved as reprint & 8 \\
Uncertain & 3 \\
P / R / F1 on resolved audit cases & 1.000 / 0.957 / 0.978 \\
\bottomrule
\end{tabular}
\end{table}

\section{Analysis}
Across methods, pair-level aggregation is the dominant signal source; the remaining difficulty lies in boundary refinement. A shallow decision tree already reaches 0.909~F1 on the main labeled slice but drops to 0.063 on hard cases. Direct LLM inference yields high recall at the cost of much broader positive outputs. The final workflow achieves the highest \texttt{all-labeled} F1 and the highest hard-case precision, but this advantage is relative and concentrated specifically in boundary control on ambiguous pairs; automated adaptation trades that precision for markedly higher hard-case recall. More broadly, this pattern shows that staged rule refinement can encode domain knowledge about coherent textual transmission in a compact and inspectable decision process.

The two target domains require different interpretations. On ECCO--ECCO, a targeted manual review provides a quantitative bound on this effect: 81 of 104 checked apparent false positives (78\%; 86\% excluding uncertain cases) were confirmed as genuine cases outside the current ground truth, indicating that strict full-corpus precision is substantially conservative. On ECCO--Newspaper, the most informative finding is that confirmed cases are rarely near-complete copies: across the 175 coverage-linked predicted positives, mean maximum coverage is approximately 0.473 (median 0.457), with only 37 cases exceeding 0.8. Many recovered items are better described as half-essay extracts or recomposed discourse units, which is what motivates an essay-scale framing instead of strict exact-copy detection.

A further property of the staged workflow is the auditability of its evidence. Each prediction is supported by inspectable signals instead of an opaque score: the coverage ratio indicates how much of the source essay is present, the bundle structure shows where evidence concentrates in the target document, and the chain structure reveals whether fragments cohere into a continuous argumentative unit. These signals carry meaning independently of the classification boundary, so each pair reaches the historian with an evidence profile that can be inspected and contested separately from the classifier's decision.

The newspaper results also reflect a property of the medium itself. Unlike books, newspapers treat essay-scale boundaries as fluid: a reused Hume essay may appear as an extract embedded within an editorial, as a recomposed digest spanning multiple columns, or as a verbatim reprint ascribed to a different source. These are historically distinct modes of transmission, so what counts as ``the same text'' is a historical question and not only a data quality problem. The 48 predicted positives that do not name Hume illustrate that transmission chains frequently exceed what bibliographic metadata can surface. The workflow also yields structured provenance hypotheses: issue-level duplication, source-side multiplicity across edition families, and edition-branch variants. Provenance structure is thus a finding in its own right, not a byproduct of detection.

\paragraph{Limitations.}
The corpus centers on a single author (Hume), limiting direct generalizability to other writers or historical periods. The Hume-centered design should therefore be read as a historically motivated stress test rather than as an author-independent benchmark. Hume is useful because the source side is well documented and widely reprinted, but authors with weaker bibliographic control, shorter works, or different quotation cultures may require different thresholds and additional expert calibration. Likewise, our comparison to Passim is downstream of fragment retrieval: the paper evaluates pair-level consolidation over retrieved hits, not replacement of newspaper-scale local matching systems. The \texttt{hard} slice overlaps with hard-case pools used during workflow refinement and carries only 22 positives, making hard-slice F1 differences statistically indistinguishable and those results diagnostic, not truly held-out. The newspaper annotation was non-blind and no inter-annotator agreement was computed, constraining the strength of precision claims on the newspaper side. The study is also not an OCR-robustness benchmark: noisier newspaper corpora may require recalibrated thresholds or additional OCR/layout-aware filtering. Finally, our LLM results cover only direct-prompt binary classification. Because Hume is canonical, these baselines are also uncontrolled external-knowledge conditions, and our results should not be read as showing that rule workflows generally outperform LLMs. A stronger future comparison would test LLMs as rerankers or rule proposers over the same candidate evidence, alongside long-context or retrieval-augmented designs.

\section{Conclusion}
We presented a workflow study of essay-scale republication and reuse detection across two structurally distinct historical environments: ECCO books, supporting quantitative evaluation over partially labeled pairs, and eighteenth-century newspapers, requiring deployment-and-audit in the absence of prior ground truth. In this candidate-reduction setting, staged pair-level evidence consolidation produced a compact, auditable candidate space while preserving useful recall; manual audit confirms all 176 newspaper predictions as genuine. Closed-world F1, deployment size, and historian audit proved to be complementary signals; under incomplete positive coverage none of them alone characterises method quality. Future work includes broader corpora and author sets to test workflow portability, multimodal page evidence for layout-dependent cases, and sequential LLM--workflow pipelines in which LLM candidate expansion feeds into staged rule refinement.

\begin{acks}
We thank the Area Chair and anonymous reviewers for their constructive feedback. This work was supported by the European Union's Horizon Europe programme through MSCA Doctoral Networks 2022 (Grant Nos. 101120349 and 101119511). We also acknowledge CSC -- IT Center for Science, Finland, for access to the LUMI supercomputer, owned by the EuroHPC Joint Undertaking and hosted by CSC and the LUMI consortium.
\end{acks}

\section*{Data Availability}
The pair-level feature data, labeled anchor splits, and workflow prediction outputs used in this paper are archived at \url{https://github.com/COMHIS/reprint-detection-hume-case-study}, which will be made publicly accessible upon publication. The underlying ECCO corpus is accessible via institutional subscription; the Burney Newspapers Collection is available through Gale Primary Sources~\cite{burney_collection}.

\section*{GenAI Disclosure}
The authors used ChatGPT and Claude to assist with writing and editing. All scientific content, experimental design, results, and conclusions are the sole responsibility of the authors.

\bibliographystyle{ACM-Reference-Format}
\balance
\bibliography{reprint_ref}

@misc{qwen3,
  author    = {{Qwen Team}},
  title     = {Qwen3 Technical Report},
  year      = {2025},
  eprint    = {2505.09388},
  archivePrefix = {arXiv},
  note      = {Model: \texttt{Qwen/Qwen3-30B-A3B-Instruct-2507}; \url{https://huggingface.co/Qwen/Qwen3-30B-A3B-Instruct-2507}}
}

@article{darnton1982history,
  author  = {Darnton, Robert},
  title   = {What Is the History of Books?},
  journal = {Daedalus},
  volume  = {111},
  number  = {3},
  pages   = {65--83},
  year    = {1982}
}

@misc{burney_collection,
  title        = {Seventeenth and Eighteenth Century {Burney} Newspapers Collection},
  author       = {{Gale}},
  howpublished = {Gale Primary Sources},
  note         = {\url{https://www.gale.com/primary-sources/seventeenth-and-eighteenth-century-burney-newspapers-collection}}
}

@inproceedings{smith2015computational,
  author    = {Smith, David A. and Cordell, Ryan and Dillon, Elizabeth Maddock and Stramp, Nick and Wilkerson, John},
  title     = {Detecting and Modeling Local Text Reuse},
  booktitle = {Proceedings of the 14th ACM/IEEE-CS Joint Conference on Digital Libraries (JCDL)},
  year      = {2014},
  pages     = {183--192},
  doi       = {10.1109/JCDL.2014.6970166}
}

@article{rosson2023reception,
  author    = {Rosson, David and M{\"a}kel{\"a}, Eetu and Vaara, Ville and Mahadevan, Ananth and Ryan, Yann and Tolonen, Mikko},
  title     = {Reception Reader: Exploring Text Reuse in Early Modern British Publications},
  journal   = {Journal of Open Humanities Data},
  year      = {2023},
  volume    = {9},
  number    = {1},
  pages     = {5},
  doi       = {10.5334/johd.101}
}

@article{bekker2020learning,
  author    = {Bekker, Jessa and Davis, Jesse},
  title     = {Learning from Positive and Unlabeled Data: A Survey},
  journal   = {Machine Learning},
  year      = {2020},
  volume    = {109},
  number    = {4},
  pages     = {719--760},
  doi       = {10.1007/s10994-020-05877-5}
}

@inproceedings{buckley2004retrieval,
  author    = {Buckley, Chris and Voorhees, Ellen M.},
  title     = {Retrieval Evaluation with Incomplete Information},
  booktitle = {Proceedings of the 27th Annual International ACM SIGIR Conference on Research and Development in Information Retrieval},
  year      = {2004},
  pages     = {25--32},
  doi       = {10.1145/1008992.1009000}
}

@article{ratner2017snorkel,
  author    = {Ratner, Alexander J. and Bach, Stephen H. and Ehrenberg, Henry and Fries, Jason and Wu, Sen and R{\'e}, Christopher},
  title     = {Snorkel: Rapid Training Data Creation with Weak Supervision},
  journal   = {Proceedings of the VLDB Endowment},
  year      = {2017},
  volume    = {11},
  number    = {3},
  pages     = {269--282},
  doi       = {10.14778/3157794.3157797}
}

@inproceedings{wu2019errudite,
  author    = {Wu, Tongshuang and Ribeiro, Marco Tulio and Heer, Jeffrey and Weld, Daniel S.},
  title     = {Errudite: Scalable, Reproducible, and Testable Error Analysis},
  booktitle = {Proceedings of the 57th Annual Meeting of the Association for Computational Linguistics (ACL)},
  year      = {2019},
  pages     = {747--763},
  doi       = {10.18653/v1/P19-1073}
}

@inproceedings{boecking2020interactive,
  author    = {Boecking, Benedikt and Neiswanger, Willie and Xing, Eric P. and Dubrawski, Artur},
  title     = {Interactive Weak Supervision: Learning Useful Heuristics for Data Labeling},
  booktitle = {Proceedings of the International Conference on Learning Representations (ICLR)},
  year      = {2021}
}

@inproceedings{galhotra2021adaptive,
  author    = {Galhotra, Sainyam and Golshan, Behzad and Tan, Wang-Chiew},
  title     = {Adaptive Rule Discovery for Labeling Text Data},
  booktitle = {Proceedings of the 2021 International Conference on Management of Data (SIGMOD)},
  year      = {2021},
  doi       = {10.1145/3448016.3457334}
}

@article{smith2024language,
  author    = {Smith, Ryan and Fries, Jason A. and Hancock, Braden and Bach, Stephen H.},
  title     = {Language Models in the Loop: Incorporating Prompting into Weak Supervision},
  journal   = {{ACM/IMS} Journal of Data Science},
  year      = {2024},
  volume    = {1},
  number    = {2},
  pages     = {1--30},
  doi       = {10.1145/3617130}
}

@inproceedings{xu2020layoutlm,
  author    = {Xu, Yiheng and Li, Minghao and Cui, Lei and Huang, Shaohan and Wei, Furu and Zhou, Ming},
  title     = {{LayoutLM}: Pre-training of Text and Layout for Document Image Understanding},
  booktitle = {Proceedings of the 26th ACM SIGKDD International Conference on Knowledge Discovery \& Data Mining},
  year      = {2020},
  pages     = {1192--1200},
  doi       = {10.1145/3394486.3403172}
}

@inproceedings{appalaraju2021docformer,
  author    = {Appalaraju, Srikar and Jasani, Bhavan and Kota, Bhargava Umesh and Xie, Yusheng and Manmatha, R.},
  title     = {{DocFormer}: End-to-End Transformer for Document Understanding},
  booktitle = {Proceedings of the IEEE/CVF International Conference on Computer Vision (ICCV)},
  year      = {2021},
  pages     = {993--1003},
  doi       = {10.1109/ICCV48922.2021.00103}
}

@article{gutehrle2022processing,
  author    = {Gutehrl{\'e}, Nicolas and Atanassova, Iana},
  title     = {Processing the Structure of Documents: Logical Layout Analysis of Historical Newspapers in {French}},
  journal   = {Journal of Data Mining \& Digital Humanities},
  year      = {2022},
  doi       = {10.46298/jdmdh.9093}
}

\end{document}